\documentclass[conference]{IEEEtran}
\IEEEoverridecommandlockouts
\usepackage{cite}
\usepackage{hyperref}
\usepackage{amsmath,amssymb,amsfonts}
\usepackage{algorithmic}
\usepackage{graphicx}
\usepackage{textcomp}
\usepackage{xcolor}
\def\BibTeX{{\rm B\kern-.05em{\sc i\kern-.025em b}\kern-.08em
    T\kern-.1667em\lower.7ex\hbox{E}\kern-.125emX}}
    
\begin{document}

\title{BanglaSarc: A Dataset for Sarcasm Detection\\
\thanks{Footer.}
}

\author{\IEEEauthorblockN{Tasnim Sakib Apon}
\IEEEauthorblockA{\textit{Computer Science and Engineering} \\
\textit{BRAC University}\\
Dhaka, Bangladesh \\
sakibapon7@gmail.com}
\and
\IEEEauthorblockN{Ramisa Anan}
\IEEEauthorblockA{\textit{Computer Science and Engineering} \\
\textit{BRAC University}\\
Dhaka, Bangladesh \\
ramisa.anan@g.bracu.ac.bd}
\and

\IEEEauthorblockN{ Elizabeth Antora Modhu}
\IEEEauthorblockA{\textit{Computer Science and Engineering} \\
\textit{BRAC University}\\
Dhaka, Bangladesh \\
elizabeth.antora.modhu@g.bracu.ac.bd}
\and
\IEEEauthorblockN{Arjun Suter}
\IEEEauthorblockA{\textit{Computer Science and Engineering} \\
\textit{BRAC University}\\
Dhaka, Bangladesh \\
arjun.suter@g.bracu.ac.bd}
\and
\IEEEauthorblockN{Ifrit Jamal Sneha}
\IEEEauthorblockA{\textit{Computer Science and Engineering} \\
\textit{BRAC University}\\
Dhaka, Bangladesh \\
Ifrit.jamal.sneha@g.bracu.ac.bd}
\and
\IEEEauthorblockN{MD. Golam Rabiul Alam}
\IEEEauthorblockA{\textit{Computer Science and Engineering} \\
\textit{BRAC University}\\
Dhaka, Bangladesh \\
rabiul.alam@bracu.ac.bd}
}

\maketitle

\begin{abstract}
    Being one of the most widely spoken language in the world, the use of Bangla has been increasing in the world of social media as well. Sarcasm is a positive statement or remark with an underlying negative motivation that is extensively employed in today's social media platforms. There has been a significant improvement in sarcasm detection in English over the previous many years, however the situation regarding Bangla sarcasm detection remains unchanged. As a result, it is still difficult to identify sarcasm in bangla, and a lack of high-quality data is a major contributing factor. This article proposes BanglaSarc, a dataset constructed specifically for bangla textual data sarcasm detection. This dataset contains of 5112 comments/status and contents collected from various online social platforms such as Facebook, YouTube, along with a few online blogs. Due to the limited amount of data collection of categorized comments in Bengali, this dataset will aid in the of study identifying sarcasm, recognizing people's emotion, detecting various types of Bengali expressions, and other domains. The dataset is publicly available at https://www.kaggle.com/datasets/sakibapon/banglasarc.
\end{abstract}


\section{Introduction}

     Sarcasm is most frequently employed to make fun of someone or something or to express discontent. It signifies the exact opposite of what we are trying to convey. Sarcasm is widely used nowadays to make fun of or show dislike, and it is ingrained in internet culture everywhere. In person interactions, online chat rooms, and comment sections of social networking sites all employ sarcasm. \par 
     
     Sarcasm detection algorithms that are automated can be useful for a variety of NLP applications, such as marketing analysis, opinion mining, and data classification \cite{Suhaimin} \cite{Moores}. Sarcasm frequently modifies the polarity of a statement, making its identification and processing critical in an automated NLP system\cite{{Wicana}}. Automation of the sarcasm detection process has received a lot of attention recently since it enables more precise analytics in online comments and reviews. Sarcasm detection in English has improved dramatically over the past several years as a result of the development of artificial intelligence (AI) technology, specifically NLP, mood and emotion analysis approaches are being used to examine predicting and eliminating phrases that indicate sarcasm. 
     However, it continues to be a barrier for the Bangla language, and only a limited amount of work has been conducted to resolve this fundamental issue. Our study's overarching objective is to introduce a sarcasm dataset that can aid in the process of automating sarcasm detection in Bangla language. The most significant challenge we encountered while performing our study was the dearth of Bangla sarcastic content on various platforms. \par
     
     In this paper, we aim to close the communication gap by developing a resource for sarcasm detection in Bangla. Our contributions are listed below:
     
    \begin{itemize}
        \item A dataset on sarcasm has been compiled from Facebook, YouTube, and other online sites in order to validate the proposed model.
        \item Data pre-processing has been used to clean the data after manual data checking.
        \item The dataset is made available to the public so that researchers can utilize it as needed.

    \end{itemize}

     In our paper, section II briefly discuss relevant work in the fields of Bangla datasets. In section III, the system model has been described. Section IV demonstrates the data evaluation. Finally, section V concludes with the goal of improving the performance of our dataset and future research.

\section{Related Works}
        
    A research by Hossain et al. present 50K annotated fake news dataset that can be used to construct automated fake news detection systems for a low-resource language like Bangla. The dataset is labeled as authentic and fake. In addition, they examine the dataset and build a benchmark system to detect Bangla fake news using state-of-the-art NLP algorithms. Moreover, they investigate classical language aspects as well as neural network-based methodologies to develop this system. This study also said that the dataset will be a beneficial resource for developing systems to combat the spread of fake news and to contribute to research with limited resources \cite{Hossain}. \par
    
    Ahmed et al. published a Bangla text dataset to ensure a safe and harassment free online environment. The information in this article was compiled and categorized from people's comments on public Facebook posts by celebrities, government figures, and athletes. There have been 44001 comments collected in total and the dataset is categories as bully and non-bully.  Furthermore, the dataset is gathered in order to train a machine to determine whether a comment is a bully expression or not, and how inappropriate it is when the comment is offensive. This study also includes exploratory analysis from several perspectives to provide a comprehensive overview \cite{Ahmed}. \par

   For instance, Ali et al. presents "BanglaSenti" a lexicon-based corpus or dataset created especially for sentiment analysis from textual data. This dataset comprises 61582 Bangla words that are positive, negative, or neutral. In order to understand the utility of this dataset, this article used not just the corpus but also a model simulation. The dataset has been standardized as the English SentiWordNet dataset so that academics can use it in the same code format. Though the dataset was created for sentiment analysis, it might also be used for emotion recognition, opinion mining, and other applications \cite{Ali}. \par
   
   Another study by Rahman et al. provides a pair of Datasets containing both Cricket and Restaurant reviews comments. For ABSA, they collected two Bangla datasets. For this dataset, positive, negative, and neutral polarities were considered. They conducted statistical linguistic analysis on the datasets and applied state-of-the-art machine learning algorithms to the acquired datasets and achieving satisfactory accuracy. The Cricket dataset has 2900 comments from online sources, whereas the Restaurant dataset contains 2800 comments. These datasets are intended to perform two tasks covering aspect category extraction and polarity identification for that aspect category \cite{Rahman}. \par
   
   Unlike the datasets mentioned above, we generated a Bangla sarcasm Dataset specifically for detecting sarcasms in bangla language. 
  
    \begin{table}[!t]
        \caption{Comparison of various studies on Datasets}
            \begin{center}
                    \begin{tabular}{|c |c |c |c |}
                        \hline
                            Dataset & Classes & Quantity  & Ref  \\
                        \hline
                            Bangla Fake News & Authentic/fake & 50000 &\cite{Hossain}\\ 
                        \hline
                            Bangla Text & Bully/Non-bully & 44001 &\cite{Ahmed}\\
                        \hline
                            BanglaSenti & Positive/Negative/ & 61582  &\cite{Ali}\\
                             & Neutral  &  & \\
                        \hline
                            Cricket and Restaurant Reviews & Positive/Negative/ & 5700  &\cite{Rahman}\\
                            & Neutral  &  & \\
                        \hline
                        
                    \end{tabular}
                    \label{fig:x RelatedWorkTable}
            \end{center}
    \end{table}

        \begin{figure*}[!t]
            \centering
            \includegraphics[scale= .9]{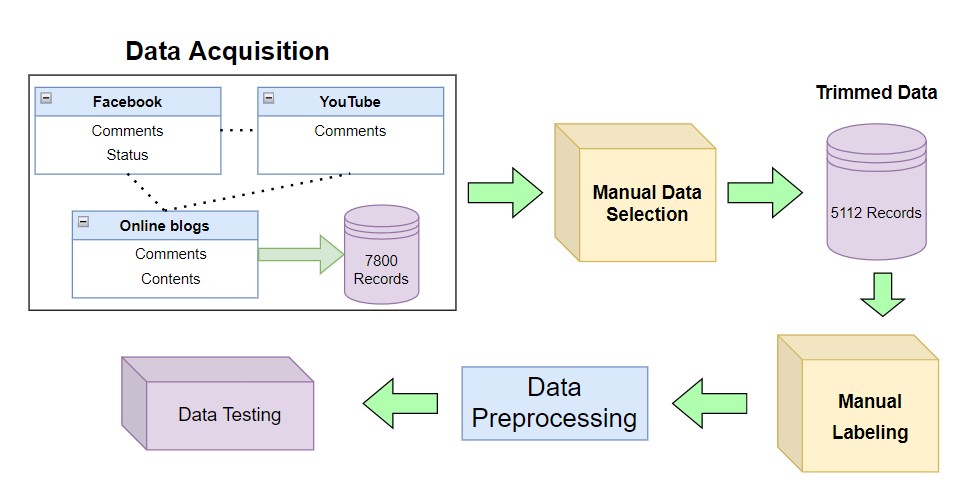}
            \caption{System Model: .}
            \label{SystemModel}
        \end{figure*}

\section{System model}

    In this section, we get an overview of our system model for sarcasm detection dataset, which is separated into four sub-sections. section \ref{DataAcquisition} of this paper talks about the data acquisition and description. However, section \ref{DataSelection} discuss the data selection part and section\ref{DataDescription} discussion the Data Description. Finally, \ref{DataPre-processing} describe the data pre-processing. \par
        
    Initially, we initiated our study by collecting comments/ status from various social media platforms along with a few online blogs. In this stage we have collected 7800 record. However, in the later stage we performed manual data selection where we manually picked the required data to produce suitable dataset for the Bangla language. In this stage we have considered a series of factors and removed roughly 2700 data. Furthermore, we label the data manually in this stage. Than we move on to data pre-processing we have considered a series of data pre-processing technique in order to offer a clean data. Lastly, we trained several machine learning models that produced the predicted results to assess the data quality. 
    
    Figure \ref{SystemModel} represents the general System Model of our study.

    \subsection{Data Acquisition} \label{DataAcquisition}
        \begin{figure}[!t]
            \centering
            \includegraphics[scale= .35]{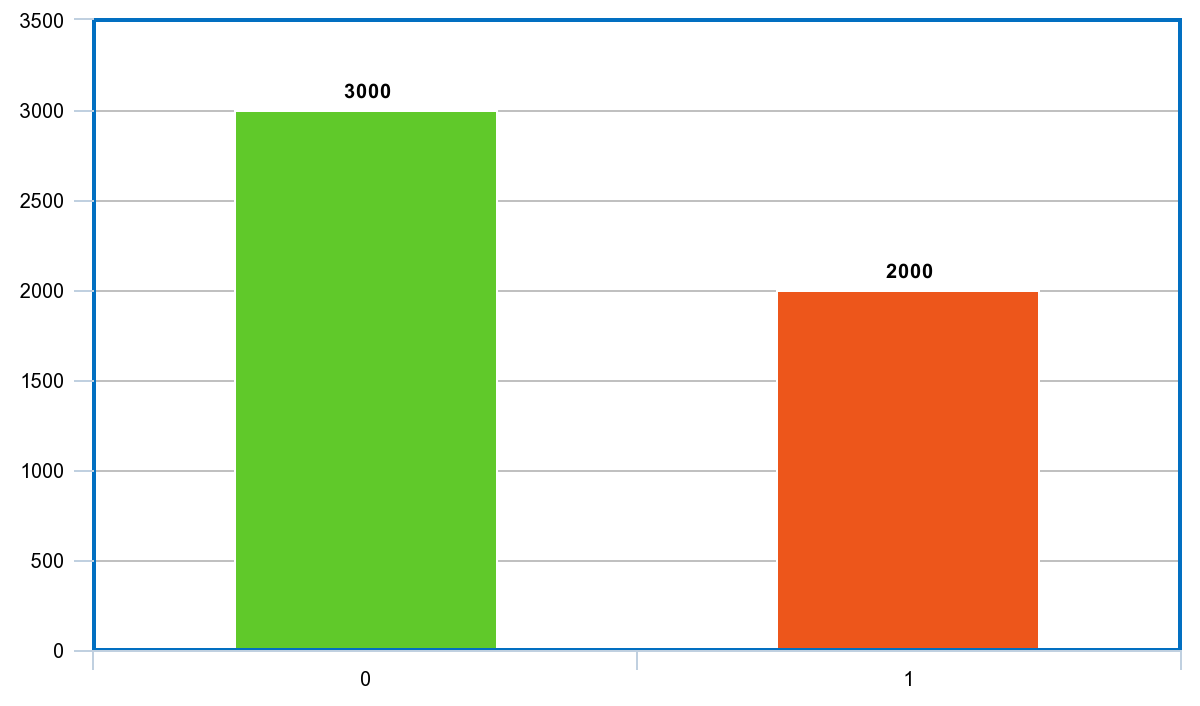}
            \caption{Distribution of Data based on class. Here, 0 represents Non-Sarcasm Data whereas 1 is represented as Sarcasm.}
            \label{DataDistribution}
        \end{figure}

        Our data acquisition presented us with a few challenges. Our initial target was various social platforms like Facebook, YouTube. However, Bangla sarcasm data is very limited, in a later period we have also considered a few online blogs. We have employed a series a data scraping technique in order to automate the process of data acquisition. Due to the fact that most websites typically restrict requests for data scraping, this technique also become a little troublesome. However, We were able to compile a Bangla sarcasm dataset comprised of comments/status from Facebook, YouTube, and various online blog contents. In our initial data acquisition period we have collected around 7800 records.\par


    \subsection{Data Selection} \label{DataSelection}
        The purpose of the step is to determine the relevant data in order to create a appropriate datasets for the Bangla language. After initial data acquisition we examined and categorized each of the assertions manually and removed noisy data throughout the screening step. Due to the Bangla language's linguistic perspective, we had to remove a large number of records for various reasons. We discovered several personal statements where users of social media mentioned their friends or provided phone numbers and other information. We have removed them from our dataset. Additionally, because the majority of our data is made up of comments gathered from multiple social media platforms, it is frequently short and meaningless. We excluded them as well. Moreover, while we conducted this screening process we also deleted repetitive data. We initially collected 7800 records, however after manually filtering, we had to remove roughly 2700 data. As a consequence, our dataset comprises 5112 comments where 3159 non-sarcastic statements, and 1951 sarcastic statements. Additionally, We  labeled the data with '0' and '1', where '0' signifies a non-sarcastic sentence and '1' denotes a sarcastic sentence in this case. 

    \subsection{Data Description} \label{DataDescription}

    Figure \ref{DataDistribution} represents the distribution of data. As mentioned previously that we have collected 3000 non sarcastic content that is depicted by 0 and 2000 sarcastic content which is represented by 1. 
    
\begin{figure*}[htbt!]
            \centering
            \includegraphics[scale= 1.55]{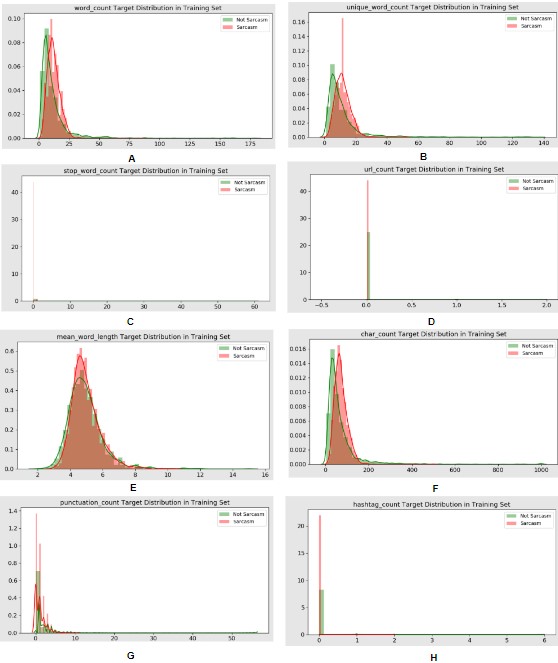}
            \caption{Distributions of meta features. Here, A represents the word Count. B denotes the Unique Word Count. C is Stop word Count. D indicates the Url Count. E is Mean Word Length. F symbolizes the Characters count. G is Punctuation Count and finally H is Hashtag count.}
            \label{DataInformation}
        \end{figure*}

    Figure \ref{DataInformation} represents the distributions of meta features in our presented dataset. In the Figure, A represents the number of words in a text. B is the number of unique words in text. Number of stop words in text is denoted in C. D is the number of URL in the text. Average Character count is presented in E. F is the number of characters present in a text. G is the number of punctuation in a text. And finally, H is the number of hashtags. 
    
    \subsection{Data Pre-processing} \label{DataPre-processing}
    
        \begin{figure}[!h]
            \centering
            \includegraphics[scale= .6]{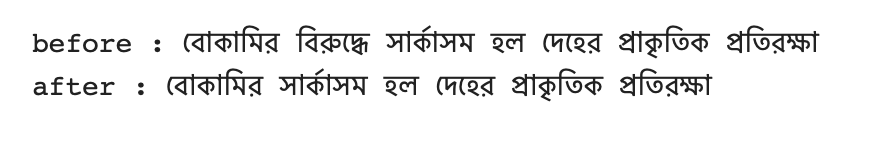}
            \caption{Data Preprocessing: Sample Input.}
            \label{Dataprocessing}
        \end{figure}
        We have considered a series of data pre-processing technique in order to clean our dataset. Although we deleted repeated information throughout the screening phase, it was done manually. Therefore, at this point, we attempted automating the removal of repeated records. In essence, we focused on offering a dataset without any noise so we removed numeric values from the data as they do not offer any value in terms of detecting sarcasm.
        Then, since handling emojis and emoticons is crucial when doing text preparation tasks in NLP, we delete all of the emoji. Once the stop words have been eliminated, the phrase will make more sense without them. They are typically removed from datasets to enhance model performance. After tokenizing the text, we eliminated all of the punctuation. A written document is tokenized when it is divided up into smaller bits or chunks called tokens. A crucial stage in the creation of a machine learning model is data preparation. Figure \ref{Dataprocessing} depicts an example of input and output produced by our preprocessing techniques.

\section{Data Evaluation}
       \begin{table}[htbt!]
            \caption{Data Performance Evaluation: Traditional Machine Learning Models}
            \begin{center}
                    \begin{tabular}{|c |c |c |c|}
                        \hline
                        Model  & Precision & Recall & F-1 Score\\ [0.5ex]
                        \hline
                        Random Forest               & 89.93  & 89.93 & 89.93 \\
                        \hline
                        Decision Tree               & 83.58  & 83.58 & 83.58 \\
                        \hline
                        K-Nearest Neighbor          & 74.78  & 74.78 & 74.78 \\
                        \hline
                        Support-Vector Machines     & 71.55 & 71.55 & 71.55 \\
                        \hline
                        Multinomial Naive Bayes     & 65.10 & 65.10 & 65.10 \\
                        \hline
                        Logistic Regression         & 62.46 & 62.46 & 62.46 \\
                        \hline
                        Stochastic Gradient Descent & 53.86 & 53.86 & 53.86 \\
                        \hline
                    \end{tabular}
                    \label{MlModels}
            \end{center}
        \end{table}

        \begin{figure*}[htbt!]
            \centering
            \includegraphics[scale= .8]{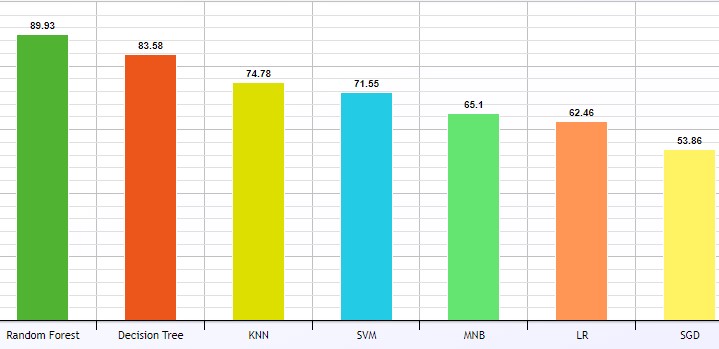}
            \caption{Traditional Machine Learning Model's Accuracy.}
            \label{MlModelAccuracy}
        \end{figure*}
    We have trained several machine learning algorithms to evaluate our dataset. These algorithms provided satisfactory outcome with higher accuracy. Additionally, several evaluation metrics including Precision, Recall, and F-1 Scores are used, and their results were too outstanding which states, sarcasm detection model can be easily trained with our presented dataset. \par
    Figure \ref{MlModelAccuracy} and Table \ref{MlModels} represents our finding after training with various machine learning algorithms on our presented dataset. Figure \ref{MlModelAccuracy} depicts the gained accuracy by each of the algorithms whereas Table \ref{MlModels} reveals the precision, recall and F-1 Score. We have utilized seven different machine learning models that are Logistic Regression, Decision Tree, Random Forest, Multinomial Naive Bayes, K-Nearest Neighbor, Kernel Support Vector Machine and Stochastic Gradient Descent. Among them, Stochastic Gradient Descent performed poorly gaining only 53.86\% accuracy. Accuracy ranging from 62.46\% to 74.78\%, Logistic Regression, Multinomial Naive Bayes, Support Vector Machine, K-nearest Neighbor's performance was moderate. However, Random Forset and Decision Tree's performance was notable. Decision Tree gained 83.58\% accuracy whereas Random Forest managed to reach 89.93\% accuracy with similar precision, recll and F-1 score. 
        
\section{Conclusion}
    The study introduces a Bangla sarcasm datasets for detecting sarcasm. A total of 5112 Bangla datasets with sarcastic and non-sarcastic notation have been gathered. This dataset was created with great care and we expect that it will have a significant influence on other researchers to detect sarcasm.

\section{My Appendix}
    Supplementary Data for this study may be accessed online  \href{https://www.kaggle.com/datasets/sakibapon/banglasarc}{here.}

\end{document}